\documentclass[11pt]{article}
\usepackage[utf8]{inputenc}
\usepackage{amsmath, amssymb, amsfonts}
\usepackage{graphicx}
\usepackage{hyperref}
\usepackage{authblk}
\usepackage{float}
\usepackage{subcaption}
\usepackage{geometry}
\title{\textbf{Learning Lie Group Generators from Trajectories}}
\author{Lifan Hu \\
\texttt{lifan.hnus@gmail.com}}
\affil{School of Computing, National University of Singapore}
\date{\today}
\begin{document}
\sloppy
\maketitle

\begin{abstract}
This work investigates the inverse problem of generator recovery in matrix Lie groups from discretized trajectories. Let $G$ be a real matrix Lie group and $\mathfrak{g} = \text{Lie}(G)$ its corresponding Lie algebra. A smooth trajectory $\gamma($t$)$ generated by a fixed Lie algebra element $\xi \in \mathfrak{g}$ follows the exponential flow $\gamma($t$) = g_0 \cdot \exp(t \xi)$. The central task addressed in this work is the reconstruction of such a latent generator $\xi$ from a discretized sequence of poses $ \{g_0, g_1, \dots, g_T\} \subset G$, sampled at uniform time intervals.

This problem is formulated as a data-driven regression from normalized sequences of discrete Lie algebra increments $\log\left(g_{t}^{-1} g_{t+1}\right)$ to the constant generator $\xi \in \mathfrak{g}$. A feedforward neural network is trained to learn this mapping across several groups, including \text{SE(2)}, \text{SE(3)}, \text{SO(3)}, and \text{SL(2,$\mathbb{R})$}. It demonstrates strong empirical accuracy under both clean and noisy conditions. This validates the viability of data-driven recovery of Lie group generators using shallow neural architectures.
\end{abstract}

\section{Introduction}
The study of continuous transformations in geometry, robots, physics, and control theory often leads natrually to the framework of Lie groups and Lie algebras. A matrix Lie group $G\subset\text{GL(n, $\mathbb{R}$)}$ is a smooth manifold endowed with a group structure, and its associated Lie algebra $\mathfrak{g} = \text{Lie}(G)$ serves as its tangent space at the identity, providing a local linearization of group behavior. The exponential map $\exp\colon \mathfrak{g} \to G$ allows for the generation of smooth one-parameter subgroups of the form $\gamma($t$) = g_0 \cdot \exp(t \xi)$, where $\xi\in\mathfrak{g}$ is a fixed generator and $g_0\in G$ is an initial condition.

This exponential flow governs a wide class of rigid-body motions, camera pose evolutions, and linear dynamic systems. In such cases, $\xi$ encapsulates the underlying motion model: angular velocity, twist, shear, or some hybrid depending on the structure of $G$. While the forward computation $\xi\mapsto\gamma(t)$ is classical, the inverse problem—recovering the constant generator $\xi$ from a sampled trajectory—is generally ill-posed under noise, discretization, and group-specific curvature effects.

This work addresses the inverse exponential problem from a data-driven perspective. Given a sequence of group elements $\{g_0, g_1, ..., g_T\} \subset G$, uniformly sampled at intervals $\Delta t$, one approximates the local velocity structure by computing ther discrete logarithmic displacements:
\[
\xi_t := \log\left(g_{t}^{-1} g_{t+1}\right) \in \mathfrak{g}.
\]
These per-step increments serve as an empirical signature of the trajectory, which is then used to regress back the original generator $\xi$. The key hypothesis is that a neural network can learn this inverse mapping from sequences $\left[\xi_0, \xi_1, \dots, \xi_{T-1}\right]$ to the latent $\xi$, exploiting regularities across sampled trajectories.

The method is evaluated on several matrix Lie groups of increasing complexity—\text{SE(2)}, \text{SE(3)}, \text{SO(3)}, and \text{SL(2,$\mathbb{R})$}—covering both Euclidean and non-Euclidean manifolds. Results show that shallow neural networks generalize well across diverse flows and noise levels, highlighting the tractability of generator recovery via learned inverse exponential maps.

This work contributes toward a computational understanding of Lie group dynamics from a learning-theoretic angle, and opens the door to using Lie-theoretic priors in geometric learning tasks.

\section{Problem Formulation}
Let $G\subset\text{GL(n, $\mathbb{R}$)}$ be a real, connected matrix Lie group, and let $\mathfrak{g}=\text{Lie}(G)\subset\mathbb{R}^{n \times n}$ denote its associated Lie algebra. Recall that $\mathfrak{g}$ is a vector space closed under the Lie bracket $[X,Y]=XY-YX$, and equipped with the exponential map $\exp\colon \mathfrak{g} \to G$, defined by the usual power series:
\[
\exp(X) = \sum_{k=0}^{\infty} \frac{1}{k!} X^k, \quad X \in \mathfrak{g}.
\]
This map is surjective in a neighborhood of the identity and provides local coordinates for $G$ around $I_n$.
Let $\xi\in\mathfrak{g}$ denote a fixed generator. The flow induced by $\xi$ defines a one-parameter subgroup $G$ given by:
\[
\gamma(t) = g_0 \cdot \exp(t \cdot \xi),\quad t \in \mathbb{R}, \quad g_0 \in G.
\]
This curve $\gamma\colon\mathbb{R} \to G$ represents a geodesic-like motion in the Lie group manifold under left-variant dynamics, and corresponds to a uniform, constant motion dictated by $\xi$.
In practice, however, continuous flows are not observed directly. Instead, one obtains a discrete sampling of the trajectory at fixed time intervals $\Delta t$, producing a sequence:
\[
\{g_0, g_1, \dots, g_T\} \subset G, \quad \text{where}\quad g_{t+1} = g_t \cdot \exp(\Delta t \cdot \xi).
\]
Assuming exact exponential integration, this implies:
\[
g_t = g_0 \cdot \exp(t \Delta t \cdot \xi), \quad \forall t \in \{0, 1, \dots, T\}
\]
Given such a sequence of group elements, the \textbf{inverse problem} is to reconstruct the original Lie algebra element $\xi \in \mathfrak{g}$ that parameterized the flow.

\section{Methodology}
After stating the \textbf{inverse problem}, this section formulates the problem precisely, introduces the learning-based approach used to approximate the inverse mapping, and outlines the data generation, architecture, and training protocols adopted to solve it.
\subsection{Trajectory Synthesis}
From the \textbf{inverse problem}: Let $G$ be a matrix Lie group and $\mathfrak{g} = \text{Lie}(G)$ its Lie algebra. A trajectory $\gamma(t) \in G$ generated by a constant $\xi \in \mathfrak{g}$ evolves via the map:
\[
\gamma(t) = g_0 \cdot \exp(t \xi), \quad t \in [0, T\cdot\Delta t].
\]
For training purposes, a dataset of such trajectories in synthesized as follows:
\begin{enumerate}
\item Sampling generators: A generator $\xi\in\mathfrak{g}\cong\mathbb{R}^n$ is randomly sampled for each trajectory within a bounded region of the Lie algebra:
\[
\xi \sim \text{Uniform}([-a,a]^n)
\]
for some fixed bound $a > 0$ appropriate to the Lie group dimension $n$. This generator is held constant across the trajectory.
\item Generation: Given an initial state $g_0 = I \in G$, the trajectory evolves by recursively applying the discrete exponential:
\[
g_{t+1} = g_t \cdot \exp(\xi \cdot \Delta t), \quad \text{for}\quad t = 0, 1, \dots, T-1
\]
where $\Delta t >0$ is a fixed timestep. This produces the trajectory $\gamma(t)$ evaluated at discrete timepoints $t\Delta t$.
\item Noise injection: To simulate real-world imprecision, zero-mean Gaussian noise $\varepsilon_k \sim \mathcal{N}(0, \sigma^2 I_n)$ is added at each timestep:
\[
g_{t+1} = g_t \cdot \exp((\xi + \varepsilon_k) \cdot \Delta t).
\]
This perturbs the motion while preserving its underlying structure, and is used to evaluate robustness during training.
\end{enumerate}

\subsection{Preprocessing and Normalization}
To convert raw trajectories into training inputs, the following steps are applied:
\begin{enumerate}
\item Discretization: The trajectory sequence is converted to local displacements:
\[
\delta_t \colon= \log\left(g_{t}^{-1} g_{t+1}\right) \in \mathfrak{g}, \quad \text{for} \quad t = 0, 1, \dots, T-1
\]
The sequence $\delta = [\delta_0, \delta_1, \dots, \delta_{T-1}] \in \mathfrak{g}^T$ represents frame-to-frame discrete displacements.

\item Dataset-wide statistics: Compute the global mean and standard deviation over all sampled $\delta_t$ values in the training dataset of size $N$:
\[
\mu = \frac{1}{N(T-1)} \sum_{i=1}^{N} \sum_{t=0}^{T-2} \delta_t^{(i)},\quad \sigma = \sqrt{ \frac{1}{N(T-1)} \sum_{i=1}^{N} \sum_{t=0}^{T-2} \left\| \delta_t^{(i)} - \mu \right\|^2 }
\]
\item Normalization: Each $\sigma_t$ is normalized using the computed statistics:
\[
\hat{\delta}_k := \frac{\delta_k - \mu}{\sigma}
\]
This produces a zero-centered, unit-scaled sequence $\hat{\delta} \in \mathbb{R}^{(T-1) \cdot \dim(\mathfrak{g})}$ for each trajectory.
\item Flattening for input: The normalized sequence is flattened to a vector form:
\[
\mathbf{x} :=  \text{vec}([\hat\delta_0,\hat\delta_1, \dots, \hat\delta_{T-1}]) \in \mathbb{R}^{(T-1) \cdot \dim(\mathfrak{g})}
\]
This vectorized input is passed to the encoder network for regression.
\end{enumerate}
\subsection{Encoder Architecture}
The goal is to learn a mapping from the normalized sequence of displacement vectors $\mathbf{x} \in \mathbb{R}^{(T-1)} \cdot \dim(\mathfrak{g})$ to the latent generator $\xi\in\mathfrak{g}$ that governs the motion via exponential flow.
This is achieved using a multilayer feedforward neural network:
\[
f_{\theta} \colon \mathbb{R}^{(T-1) \cdot \dim(\mathfrak{g})} \to \mathfrak{g},
\]
parametrized by weights $\theta$. The model consists of an input flattening layer, two hidden layers with ReLU activations, and a final linear output layer yielding a vector in $\mathbb{R}^{\dim(\mathfrak{g})}$. The network computes:
\[
f_{\theta}(x) = W_3 \cdot \rho\left(W_2 \cdot \rho\left(W_1 \cdot x + b_1\right) + b_2\right) + b_3
\]
where $\rho(\cdot)$ is the ReLU function and $W_i, b_i$ are learned weights and biases.

The network is trained to minimize the squared Euclidean loss:
\[
\mathcal{L}_{\mathrm{MSE}}(\theta) = \left\| f_{\theta}(x) - \xi \right\|_2^2
\]
This architecture is intentionally simple to emphasize geometric information in the input rather than complex temporal modeling. No recurrence, convolution, or explicit sequence bias is introduced. The model learns to regress the constant generator purely from normalized Lie algebraic displacements.
\subsection{Lie Group Variants}
The general training pipeline is applied across multiple matrix Lie groups to evaluate the encoder’s capacity to generalize over differing geometric structures. Each group $G \subset GL(n,\mathbb{R})$ comes with a corresponding Lie algebra $\mathfrak{g}$, a group exponential map $\exp\colon \mathfrak{g} \to G$ and logarithm $\log\colon G \to \mathfrak{g}$ implemented in closed form or via numerical routines.

The following groups are considered:
\begin{itemize}
\item Special Euclidean Group in 2D:
$\mathrm{SE}(2) = \mathbb{R}^2 \rtimes \mathrm{SO}(2)$, with $\dim(\mathfrak{se}(2)) = 3$. \\
Group elements are $3\times3$ matrices of the form:
\[
g =
\begin{bmatrix}
R & t \\
0 & 1
\end{bmatrix}, \quad
R \in \mathrm{SO}(2), \; t \in \mathbb{R}^2.
\]
\item Special Euclidean Group in 3D:
$\mathrm{SE}(2) = \mathbb{R}^3 \rtimes \mathrm{SO}(3)$, with $\dim(\mathfrak{se}(3)) = 6$. \\
Group elements are $3\times3$ matrices of the form:
\[
g =
\begin{bmatrix}
R & t \\
0 & 1
\end{bmatrix}, \quad
R \in \mathrm{SO}(3), \; t \in \mathbb{R}^3.
\]
\item Special Orthogonal Group:
$\mathrm{SO}(3) \subset \mathrm{GL}(3, \mathbb{R}), \quad \text{the group of pure 3D rotations.}$, the group of pure 3D rotations.
Here, $\mathfrak{so}(3) \cong \mathbb{R}^3$ via the hat map, and only angular components are modeled.
\item Special Linear Group:
$\mathrm{SL}(2, \mathbb{R}) = \left\{ A \in \mathrm{GL}(2, \mathbb{R}) \mid \det A = 1 \right\}$, a non-compact Lie group of dimension 3.
Its Lie algebra is $\mathfrak{sl}(2, \mathbb{R}) = \left\{ A \in \mathbb{R}^{2 \times 2} \mid \mathrm{tr}(A) = 0 \right\}$.

\end{itemize}
Each group introduces distinct curvature, nonlinearity, and numerical behavior in the exponential and logarithmic maps. Training on all variants uses the same encoder architecture and loss function, highlighting the group-agnostic nature of the representation learning method.

\section{Experiments and Results}
\subsection{SE(2) and SE(3): Translation and Rotation Flows}
In the SE(2) and SE(3) experiments, the model learns to recover both translational and rotational components from trajectories simulated using fixed spatial twists. Despite the low capacity of the network (two hidden layers), training converges quickly, and prediction errors are consistently below 0.03 across all dimensions. Notably, SE(3) experiments remain robust under mild Gaussian noise injection.

\begin{figure}[H]
    \centering
    \begin{subfigure}{0.28\linewidth}
        \centering
        \includegraphics[width=\linewidth]{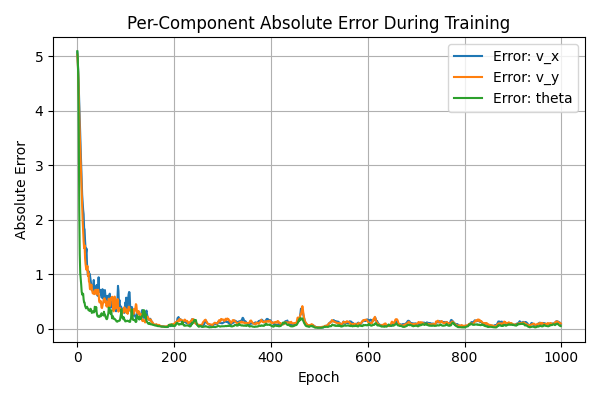}
        \caption{Per-component error trend}
    \end{subfigure}
    \hfill
    \begin{subfigure}{0.28\linewidth}
        \centering
        \includegraphics[width=\linewidth]{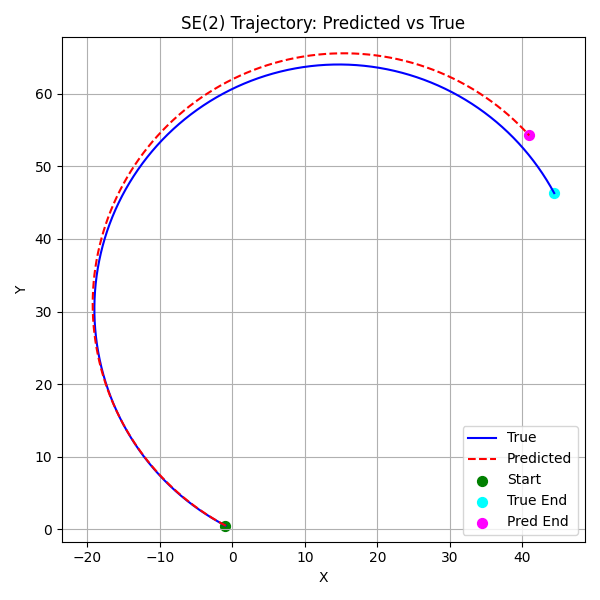}
        \caption{Predicted vs. Ground truth trajectory}
    \end{subfigure}
    \hfill
    \begin{subfigure}{0.28\linewidth}
        \centering
        \includegraphics[width=\linewidth]{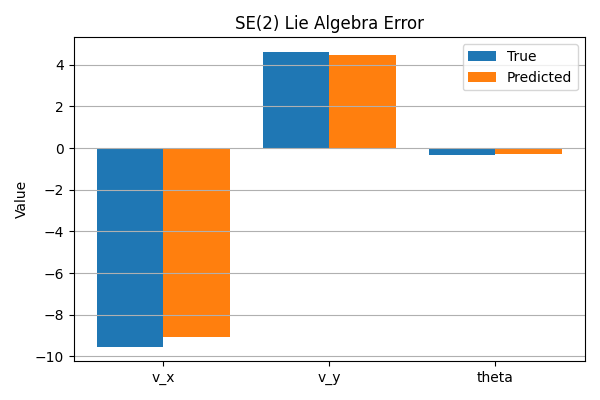}
        \caption{Estimated vs. true generator $\xi$ in Lie algebra}
    \end{subfigure}
    \caption{MSE Loss, Trajectory Comparison and of predicted and ground truth trajectories in $\mathrm{SE}(2)$.}
\end{figure}

\begin{figure}[H]
    \centering
    \begin{subfigure}{0.28\linewidth}
        \centering
        \includegraphics[width=\linewidth]{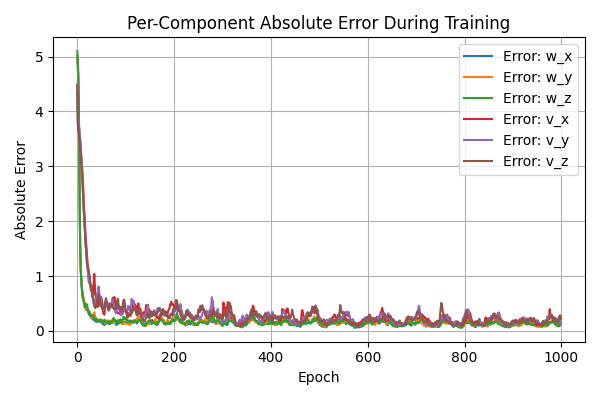}
        \caption{Per-component error trend}
    \end{subfigure}
    \hfill
    \begin{subfigure}{0.28\linewidth}
        \centering
        \includegraphics[width=\linewidth]{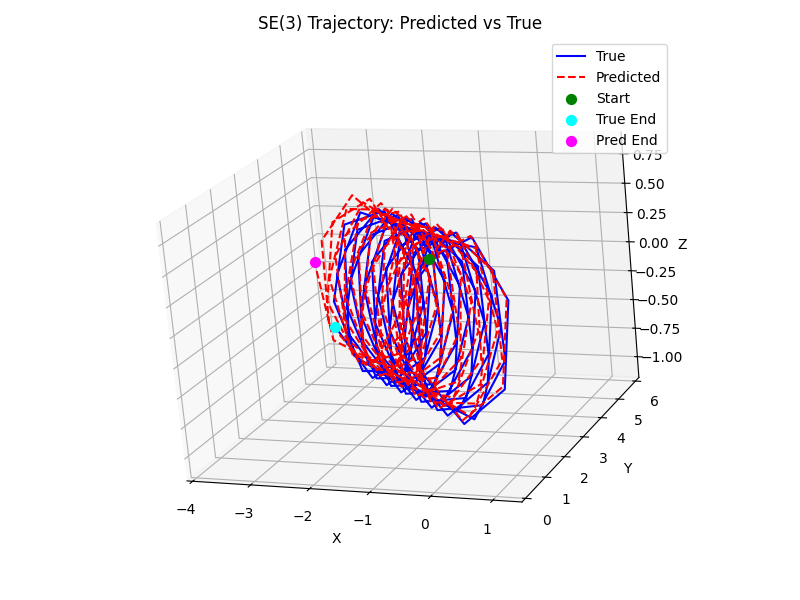}
        \caption{Predicted vs. Ground truth trajectory}
    \end{subfigure}
    \hfill
    \begin{subfigure}{0.28\linewidth}
        \centering
        \includegraphics[width=\linewidth]{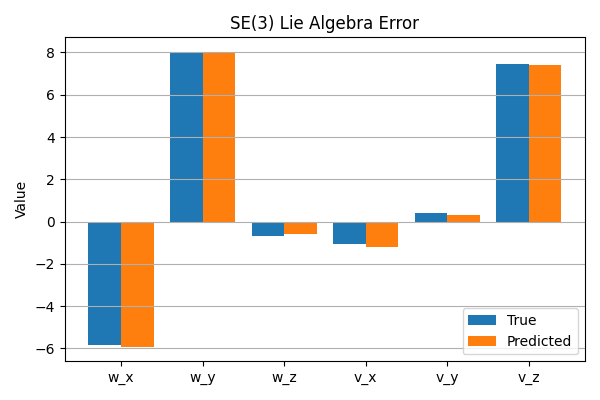}
        \caption{Estimated vs. true generator $\xi$ in Lie algebra}
    \end{subfigure}
    \caption{MSE Loss, Trajectory Comparison and of predicted and ground truth trajectories in $\mathrm{SE}(3)$.}
\end{figure}

\subsection{SO(3): Pure Rotation Learning}
In the SO(3) setting, the model is tasked with recovering a 3D angular velocity vector from sequences of relative rotations. The predicted generator reconstructs the rotation trajectory with high accuracy. Even under large angular velocities (up to 10 rad/s), orientation tracking of the z-axis direction remains visually and numerically stable.

\begin{figure}[H]
    \centering
    \begin{subfigure}{0.28\linewidth}
        \centering
        \includegraphics[width=\linewidth]{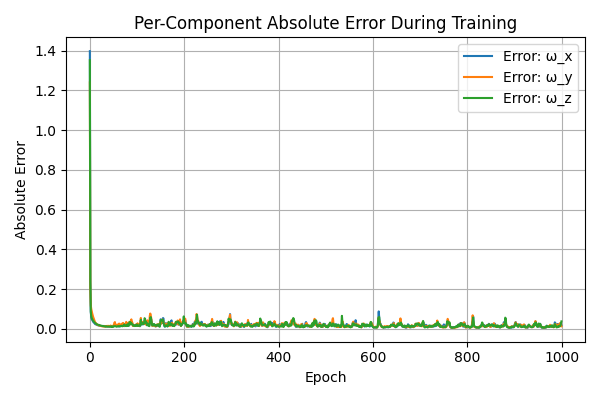}
        \caption{Per-component error trend}
    \end{subfigure}
    \hfill
    \begin{subfigure}{0.28\linewidth}
        \centering
        \includegraphics[width=\linewidth]{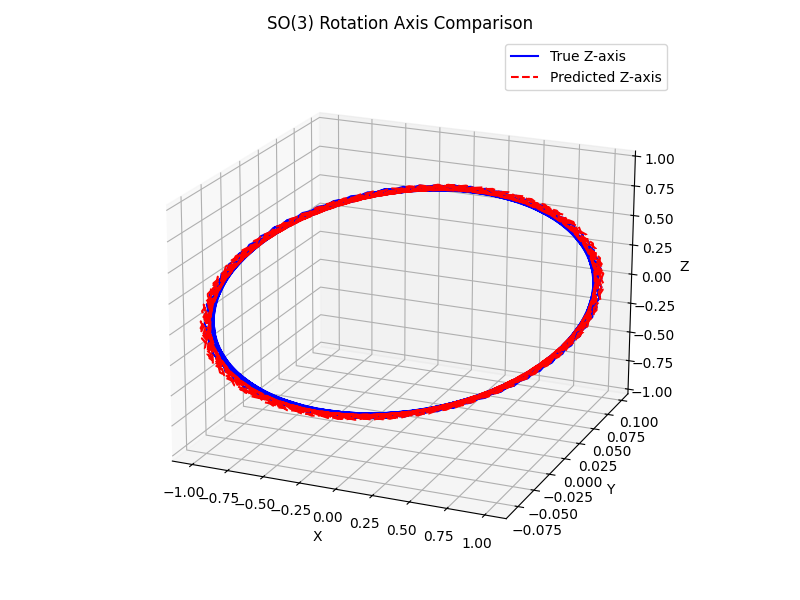}
        \caption{Predicted vs. Ground truth trajectory}
    \end{subfigure}
    \hfill
    \begin{subfigure}{0.28\linewidth}
        \centering
        \includegraphics[width=\linewidth]{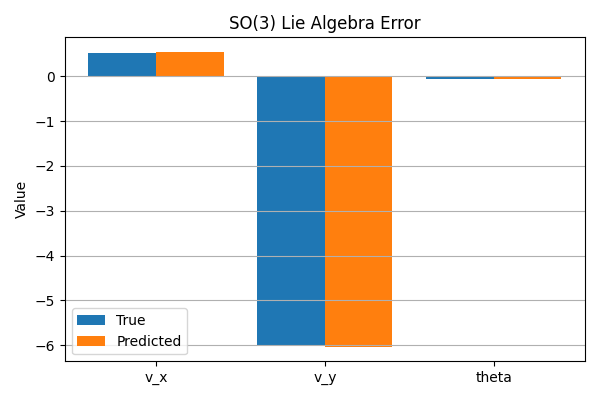}
        \caption{Estimated vs. true generator $\xi$ in Lie algebra}
    \end{subfigure}
    \caption{MSE Loss, Trajectory Comparison and of predicted and ground truth trajectories in $\mathrm{SO}(3)$.}
\end{figure}

\subsection{\texorpdfstring{$\mathrm{SL}(2,\mathbb{R})$: Dynamic Generator Recovery}{SL(2,R): Dynamic Generator Recovery}}
Training on $\mathrm{SL}(2,\mathbb{R})$ trajectories presents additional numerical challenges due to sensitivity to determinant drift and matrix inversion under floating-point noise. Nevertheless, a modified version of the encoder trained on stabilized sequences achieved reliable generator predictions. The recovered generators reflect the correct dynamical regime—elliptic, hyperbolic, or parabolic—and consistently match the ground truth within acceptable error margins.

\begin{figure}[H]
    \centering
    \begin{subfigure}{0.28\linewidth}
        \centering
        \includegraphics[width=\linewidth]{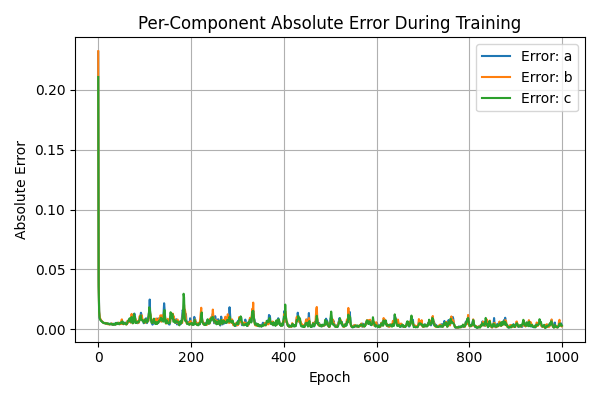}
        \caption{Per-component error trend}
    \end{subfigure}
    \hfill
    \begin{subfigure}{0.28\linewidth}
        \centering
        \includegraphics[width=\linewidth]{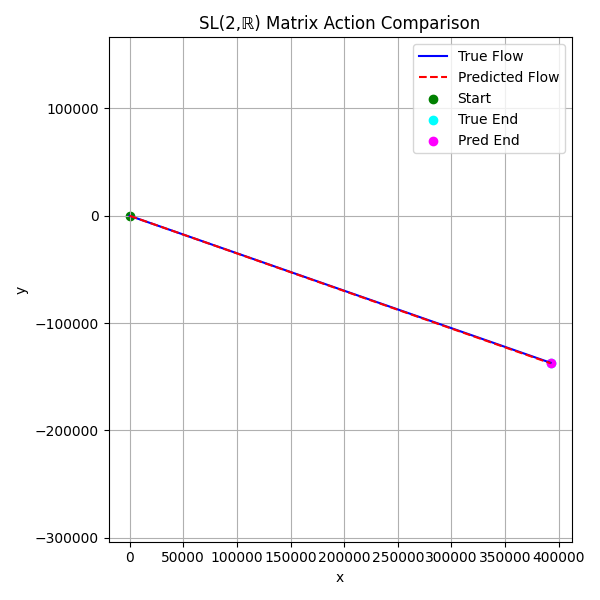}
        \caption{Matrix Flow Comparison}
    \end{subfigure}
    \hfill
    \begin{subfigure}{0.28\linewidth}
        \centering
        \includegraphics[width=\linewidth]{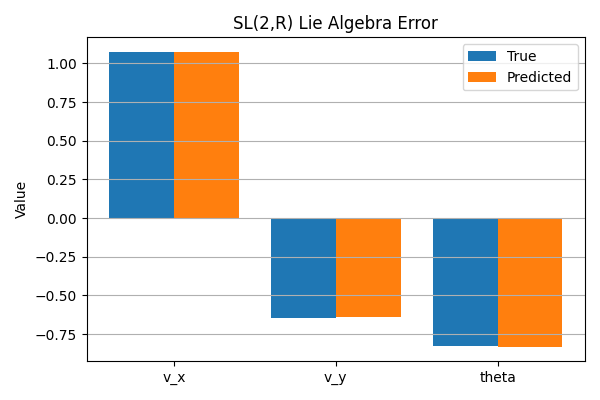}
        \caption{Estimated vs. true generator $\xi$ in Lie algebra}
    \end{subfigure}
    \caption{MSE Loss, Trajectory Comparison and of predicted and ground truth trajectories in $SL(2,\mathbb{R})$.}
\end{figure}

\section{Discussion}
The empirical results demonstrate that generator recovery via shallow neural encoders is feasible across a range of Lie groups, from compact groups like $\mathrm{SO}(3)$ to non-compact and sensitive groups such as $\mathrm{SL}(2,\mathbb{R})$. A consistent trend observed across all experiments is the effectiveness of temporal aggregation of normalized Lie algebra displacements as input features. This representation encodes geometric motion information in a group-invariant and expressive manner, reducing the learning problem to a regularized inverse map.

Despite the conceptual simplicity of the network architectures used, the models are able to capture the geometric structure of the group trajectories and converge to accurate approximations of the latent generator $\xi \in \mathfrak{g}$. Notably, no inductive bias was introduced to encode group structure, suggesting that even vanilla MLPs can serve as functional approximators for inverse exponential flows given sufficient trajectory information.

However, several limitations merit further attention. For instance, groups with sensitive determinants or non-Euclidean topology (e.g., $\mathrm{SL}(2,\mathbb{R})$) require stabilization techniques to avoid numerical singularities. In these cases, trajectory normalization and matrix regularization become essential preprocessing steps. Moreover, while the models generalize well under moderate noise, performance may degrade with increasing variance or under drift in $\Delta t$.

From a theoretical perspective, this study raises questions about the function class complexity needed to invert exponential maps across different Lie algebras. The success of shallow models implies that, at least for constant generators and well-behaved time steps, the inversion problem lies within a relatively low-complexity regime. Understanding this from a representation learning or differential geometry standpoint would be a fruitful direction for future research.

\section{Conclusion and Future Work}
This study introduced a neural framework for solving the inverse generator problem on matrix Lie groups by leveraging synthetic trajectory data and normalized algebraic displacements. By discretizing smooth exponential flows $\gamma(t) = g_0 \cdot \exp(t\xi)$ into sequences of group elements and processing their inter-step increments, the model successfully recovers the latent Lie algebra element $\xi \in \mathfrak{g}$ that governs the motion. Across diverse groups including $\mathrm{SE}(2)$, $\mathrm{SE}(3)$, $\mathrm{SO}(3)$, and $\mathrm{SL}(2,\mathbb{R})$, the results confirm that shallow neural networks can approximate the inverse Lie exponential map with competitive precision. Notably, even in sensitive regimes like $\mathrm{SL}(2,\mathbb{R})$, the recovered generators aligned with the correct geometric class (elliptic, hyperbolic, or parabolic), showcasing the robustness of this method under moderate noise.

My inspiration stems from this \cite{Joan}. To support further experimentation and reproducibility, this paper is accompanied by a complete GitHub repository containing all training scripts, Lie group definitions, and visualization utilities. Users may extend the current architecture, add new Lie groups, or integrate the encoders into broader applications in robotics, geometric control, or manifold learning. Future work may explore extensions to time-varying generators, probabilistic generator inference, or learning directly from observed sensor trajectories in real-world systems. This project may serve both as a mathematical contribution and a practical toolkit for geometric machine learning researchers and developers.

\bibliographystyle{plain}

\end{document}